\documentclass[conference]{IEEEtran}

\usepackage[noadjust]{cite}
\usepackage{multicol}

\ifCLASSINFOpdf
  \usepackage[pdftex]{graphicx}
  \graphicspath{ {./imgs/}}

\usepackage{amsmath,amsfonts}
\usepackage{algorithm}
\usepackage[noend]{algpseudocode}
\usepackage{array}
\usepackage{textcomp}
\usepackage{stfloats}
\usepackage{url}
\usepackage{verbatim}

\usepackage{subfigure, graphicx,url,times,multirow,amsmath,amssymb,algorithm,xspace,epsfig,todonotes,array,caption,color}

\DeclareMathOperator*{\argmax}{argmax}

\usepackage{float}

\begin{document}

\title{Task Relabelling for Multi-task Transfer \\using Successor Features}

\author{
\IEEEauthorblockN{Martin Balla}
\IEEEauthorblockA{Queen Mary University of London\\
m.balla@qmul.ac.uk}

\and
\IEEEauthorblockN{Diego Perez-Liebana}
\IEEEauthorblockA{Queen Mary University of London\\
diego.perez@qmul.ac.uk}}



\maketitle

\begin{abstract}
Deep Reinforcement Learning has been very successful recently with various works on complex domains. Most works are concerned with learning a single policy that solves the target task, but is fixed in the sense that if the environment changes the agent is unable to adapt to it. Successor Features (SFs) proposes a mechanism that allows learning policies that are not tied to any particular reward function. In this work we investigate how SFs may be pre-trained without observing any reward in a custom environment that features resource collection, traps and crafting. After pre-training we expose the SF agents to various target tasks and see how well they can transfer to new tasks. Transferring is done without any further training on the SF agents, instead just by providing a task vector. For training the SFs we propose a task relabelling method which greatly improves the agent's performance. 
\end{abstract}


\section{Introduction}
The combination of Reinforcement Learning (RL) with Deep Neural Networks have shown great progress on many challenging domains~\cite{mnih2015human, silver2016mastering, vinyals2019grandmaster}. Most of the success comes from problems where the goal is clearly defined, such as achieving a high-score. Many state-of-the-art work require a large amount of compute just to produce an agent that performs well on a particular task with little concern about potential changes at test time. Most works assume that the testing domain is the same as the training. Small changes to the test environment can completely break the agent's policy and fine tuning is often more challenging than retraining the agent from scratch.

Generalisation~\cite{kirk2021survey} is important for using Reinforcement Learning agents for most real-world applications. Ideally, we would like to develop agents that can quickly adapt to changes in the environment with little or no further training. In this work we are interested in a specific type of transfer where only the task (reward function) changes in the environment; the states, observations and dynamics of the environment remain unchanged. Each reward function induces a new RL objective that requires the agent to change its policy. One approach to this transfer problem is to use Goal Conditioned RL where the agent gets additional information about the objective of the task. Defining the goal signal is not trivial and often leads to generalisation problems when a previously unseen goal signal is provided to the agent. Successor Features (SFs) learn what state features (events) the agent expects to visit under its policy. Goals can be expressed in the form of task vectors that define the importance of the features. Another way to look at SFs is that the agent predicts the availability of an event and the task vector defines the desirability of such events~\cite{barreto2016successor}. 

There are many cases where transferring to new tasks is required within an environment. One of the main use cases is during game development. As the game is developed RL agents may be used to automatically find bugs~\cite{bergdahl2020augmenting, guerrero2018using}. For game testing an agent should be able to quickly evaluate changes in the game without running a full training from scratch for each change. Another use-case is using RL for Non-Player Characters (NPCs). Some games are open-ended without a clear objective (Minecraft~\cite{johnson2016malmo}). Designing a reward function that covers the behaviours that we expect from an agent is not trivial. Reward shaping is commonly used to shape the agent's policy, but it requires a lot of effort from humans to find the right reward function. Evaluating each reward function requires training an agent which might take days. Using SFs the agent can quickly adapt its policy to new reward functions.


Our contribution is the proposal of two novel task relabelling techniques for training SFs. We compare how well SFs can transfer to new tasks with different pre-training configurations and their performance when they get trained directly on the target task. Transferring to new tasks is done zero-shot, without any further training, just by providing the agents the task vector for the target task. 

The rest of the paper gives a short introduction to Reinforcement Learning and Successor Features in Section~\ref{sec:background}, followed by some related work in Section~\ref{sec:related}. Section~\ref{sec:method} explains the methods used in this paper. Section~\ref{sec:experiments} describes the types of reward functions and the environment that we use in this paper. The pre-training experiments are shown in Section~\ref{sec:pretraining} followed by the target training experiments (Section~\ref{sec:target}) and the transfer experiments (Section~\ref{sec:transfer}).
Finally, we give a conclusion and directions for future work in Section~\ref{sec:conclusion}.

\section{Background} \label{sec:background}
\subsection{Reinforcement Learning} \label{ssec:rl}
Reinforcement Learning~\cite{sutton2018reinforcement} is often modelled as a Markov Decision Process (MDP). An MDP is represented as a tuple of $M = (S, A, R, T, p)$, where $S$ is the state space, $A$ is the action space, $R$ is the reward function, $T$ is the transition function and $p$ is the set of initial states. The agent's objective is to learn a policy $\pi: S \rightarrow A$, a mapping from state to action that maximises the return $\sum_t^\infty \gamma r_t$ (cumulative discounted reward) given a discount factor $\gamma \in [0, 1]$ when following the policy.

Similarly to Barreto et al.'s~\cite{barreto2016successor} setup, instead of working with a single MDP we are using a set of MDPs $M \in \{M_1, M_2, ... M_n\}$. All MDPs share the same dynamics and visuals, only their reward function differs.  A task in this work is defined by a particular reward function (i.e: \textit{task: "collect wood" gives $1$ reward when a wood is collected by the agent $0$ otherwise}). An objective of this work is to transfer knowledge between tasks, for which we propose to have a pre-training phase where no rewards observed ($r(s) == 0; \forall s \in S$). The pre-training phase can be modelled by an MDP where the reward is zero for all states $r(s) = 0 \forall s \in S$.

\subsection{Successor Features} \label{ssec:sf}
Successor Features (SFs) are based on Successor Representation (SR), introduced by Dayan~\cite{dayan_improving_1993}. SR decomposes the state-action value function into an expected future state-occupancy $M(s, s')$ and a reward weight vector $w$. The state-occupancy function is independent from the reward, but it depends on the policy that was used to collect the data. The weight vector determines the importance of the future states. The linear combination of these factors results in the $Q$ values:
\begin{equation}
    Q(s, a) = M(s, s')^T w(s)
    \label{eq:sr}
\end{equation}

SR facilitates transfer to new tasks, just by supplying the right task vector $w$. As the SR predicts the successor states from the current state, it quickly becomes infeasible to use it for complex environments. Barreto et al.~\cite{barreto2016successor} proposed to use instead state features $\phi$, and since in most practical cases there are only a few states where $R(s) != 0$, it makes more sense to only model the state features with $\phi$ that may be rewarding. To use SFs, $\phi$ needs to extract the potentially rewarding states so that we can recover the one-step reward function as the linear combination of the state features and the task weight vector:
\begin{equation}
    R(s_t) = \phi_{s_t}^T w
    \label{eq:one_step_reward}
\end{equation}

The Successor Feature is the expected discounted future state features: 
\begin{equation}
    \psi(s, a) = E^\pi [\sum_t^\infty \gamma^t \phi(s_t)]
    \label{eq:sf}
\end{equation}

The Q function can be recovered as the linear combination of the SF and the task vector:
\begin{equation}
    Q(s, a) = \psi(s, a)^T w
    \label{eq:q_function}
\end{equation}

Barreto et al.~\cite{barreto2016successor} have noted that a trained SF agent can be quickly evaluated on any task vector just by taking the linear combination of the predicted SFs with the task vector $\psi(s, a)^T w$, which they refer to as Generalised Policy Evaluation (GPE). GPE is show in Equation~\ref{eq:gpe} where $W$ is the set of task vectors. As the SF depend on the policy used to collect the experience,~\cite{barreto2016successor} also proposed to keep multiple policies and use the Generalised Policy Improvement (GPI) to predict all SFs for all policies and pick the action that results in the highest Q value when combined with the current task vector $w$. Equation~\ref{eq:gpi} shows the GPI process where $n$ represents the number of policies.

\begin{equation} \label{eq:gpe}
    Q(s, a| w) = \psi(s, a)^T w_i ; \forall w_i \in W
\end{equation}

\begin{equation} \label{eq:gpi}
    a = \argmax_a \max_n \psi_{\pi^n}(s, a)^T w
\end{equation}

To learn the SF we may use Temporal Difference Learning:
\begin{equation}
    L_{SF}(\psi | \theta) = E[\phi_{s_t} + \gamma \psi(\phi_{s_{t+1}}, a') - \psi(\phi_{s_{t}}, a_t)]
    \label{eq:sf_loss}
\end{equation}

where $a' = argmax_a \psi(\phi_{s_{t+1}}, a)^T w$ and $\theta$ represents the parameters of the Neural Network.
To train the reward predictor we can use Supervised Learning:
\begin{equation}
    L_{rew}(w) = (R(s_t) - \phi_{s_t}^T w)^2
    \label{eq:reward_loss}
\end{equation}

Transferring to new tasks may be done without any further training just by giving the agent the right task vector $w$ (zero-shot). If the task vector is not known then it can be learnt by fixing the SF's parameters and minimising the reward prediction loss. Learning the task vector is much faster than relearning the policy as it is a Supervised Learning problem, observing a few rewards should be enough to learn the true task vector (few-shot learning).

\section{Related Work} \label{sec:related}
Dayan~\cite{dayan_improving_1993} proposed the idea of Successor Representation for transfer by breaking down the Q function into an expected future state occupancy and a reward weight function. Kulkarni et al.~\cite{kulkarni2016deep} proposed a method to approximate the SR using Neural Networks. Later Barreto et al.~\cite{barreto2016successor} noted that there is no point representing all states if only a small subset of states are rewarding and proposed to use state-features $\phi$ instead.

The Option-Keyboard~\cite{barreto2019option} combines a Hierarchical RL architecture with SFs, where SFs are used to express low-level policies and a controller is trained on top that selects task vectors as actions (action abstraction). VISR~\cite{hansen2020fast} learns the state features $\phi$ and the SF $\psi$ together in an unsupervised way based on the behavioural mutual information. Filos et al.~\cite{filos2021psiphi} combine SFs with Inverse Reinforcement Learning to learn the SFs from a dataset of various demonstrators. Machado et al~\cite{machado2020count} proposed to use the magnitude of the SF for exploration. There is evidence that the human brain might work similarly to the Successor Representation~\cite{momennejad2017successor}

Transferring in this setup is related to Goal-conditioned RL such as Universal Value Function Approximators (UVFAs)~\cite{schaul2015universal} and Hindsight Experience Replay (HER)~\cite{andrychowicz2017hindsight}). Borsa et al.~\cite{borsa2018universal} have proposed to combine UVFAs and SFs as both methods provide different generalisation capabilities. In the goal-conditioned RL setup the goal is pre-defined and transferring to new goals is challenging as they are outside the agent's distribution. Many works use goals that are a subset of the state space, which might be ambiguous in many environments (for example: when the task is to chop a tree, after completing it, the final observation might not show the tree at all, so how does the agent know what the goal is?). In this work we used goal conditioned RL agents as baselines for the settings where the goal changed between episodes.
The relabelling techniques proposed in this paper are inspired by the way HER relabels goals during training, but in our case relabelling is done on the task vector. SFs may be considered as goal conditioned agents where the goal is expressed in the form of a task vector that defines the desirability of state-features.

\section{Method} \label{sec:method}
When learning the Successor Features there are a few design decisions to make, which are discussed in this section. 

\subsection{Pre-training}
The SF depends on the policy that was used to collect the experience as it predicts the expected discounted future state features under the behaviour policy. In this sense the SF depends on the task it was trained on. If the target tasks are too different from the training tasks then we cannot expect to successfully transfer. To improve the possibility of successful transfer as pre-training we propose to try to maximise each feature individually instead of training on a set of training tasks. Since $\phi^d$ is a d-dimensional vector we can construct $d$ pre-training tasks by constructing one-hot vectors where each task represents a dimension in the feature space.


For example if the feature space is 2-dimensional, feature 1 is picking up item A and feature 2 is picking up item B, we can construct 2 pre-training tasks: 1, Collect A by setting the task $[1, 0]$ and collect B as $[0, 1]$. An advantage of pre-training in this form is that it is not limited to only collecting items: any "event" can be defined as a feature.

\subsection{Number of policies}
Using a single policy might be insufficient to cover many tasks. If an SF agent is trained in a single task, it might be unable to transfer to tasks where the gap between the target task and the training tasks is large. Using multiple policies allows us to learn a larger set of SFs. If a new task is presented it may be more relevant to one of the policies that we have trained already. Using SFs we can also apply GPI + GPE to decide which policy is the most relevant to the current task.

The approach that we follow in this work is to use as many policies as the number of state features $n = |\phi|$, which allows us to train a policy to maximise the expectation over each feature. Using many policies allows us to apply GPI (choose the most relevant policy to the task) and GPE (quickly evaluate the task vectors) as done in previous works~\cite{barreto2020fast}.

In practice, 
we could use separate networks to count on a set of policies, but this would become 
impractical as the number of features $|\phi|$ increases. Instead, we use a shared network that outputs the predictions for each policy ($n \times |A|$ output). In this work we only use either $n$ policies or $1$, but in practice one could use an arbitrary number of policies, but during training a clear objective (i.e: a target task that the policies is trained to accomplish) should be set for all of them. 


\subsection{Task Relabelling} \label{ssec:relabelling}
During training the agent is given a task in the form of a task vector $w$ and tries to maximise the occurrence of the features relevant to that task. In many cases, especially early in the training process, the agent might observe state features $\phi$ that are not relevant to the current task, but are useful to others. We propose two methods that reuse these observed features during training in order to improve the SF's performance:

\subsubsection{Hindsight Task Replacement (HTR)}

Similarly to Hindsight Experience Replay~\cite{andrychowicz2017hindsight}, we can replace these task vectors as if the observed features were the task. When sampling from the Replay Memory instead of taking the task vector $w$ that was used to collect the experience, we search for the next event the agent has observed ($\max (\phi) \neq 0)$ and use it as the task vector. If the episode ends before finding such feature then we use the original task used for the episode.

Events in this setup are one-hot vectors as events are mutually exclusive (only one event may happen in a single step). As both state features and task vectors are one-hot vectors during pre-training, we can use the state features as the task vector: $w \leftarrow \phi(s_t)$.

\subsubsection{Task Replacement (TR)}
When training SFs with more than $1$ policy we can set different objectives for them. As we are interested in pre-training agents that do well in all possible target tasks, one option is to train as many policies as the dimension of our state-features $n = |\phi|$ and make each policy responsible to maximise the occurrence of a feature.
During training, acting with respect to a particular task often leads to observing features irrelevant to the current task. In this case we can use that trajectory to not just update the current policy but all of them. For this setting we propose to relabel the task vectors during training to the policy's objective and update all of them with the sampled mini-batches.
Using multiple policies also allows the use of GPE+GPI during evaluation which might help in transferring to new tasks.


When optimising the SF we can replace the task vector $w$ that was used during the experience collection by the policy's objective. Since we use as many policies as features ($n == |\phi|$) we can replace the task vector by a one-hot version of the policy's objective. 





\section{Experiments} \label{sec:experiments}

\subsection{MiniCrafter} \label{ssec:env}
\begin{figure}[!t]
\centering
\includegraphics[scale=0.3]{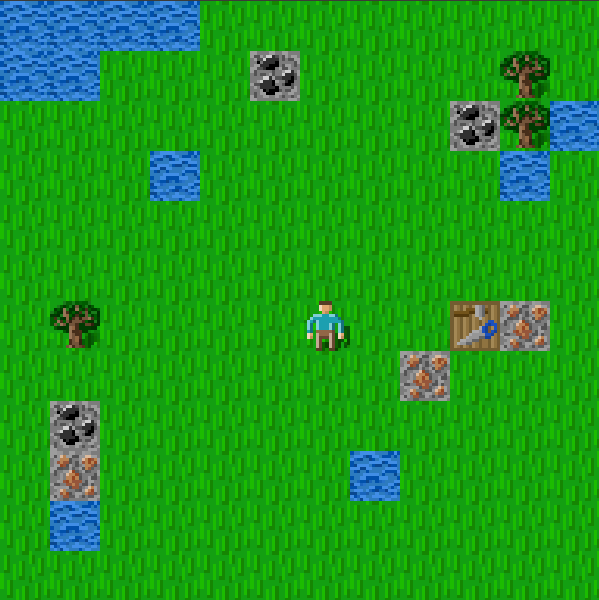}
\caption{MiniCrafter environment: Example state from the environment. The water represents the traps, the $3$ resource types wood, iron, coal are present and a single crafting table.}
\label{fig:minicrafter}
\end{figure}

\begin{figure}[!t]
\centering
\includegraphics[scale=0.35]{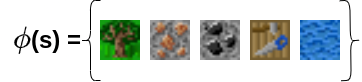}
\caption{The state features $\phi$ used in this work. An example task vector $w$ = $[0.5, 0, 0, 1, -1]$ (used in the zero-shot transfer experiments for craft\_staff). The example task vector means that the agent gets $0.5$ reward for collecting a tree, $1$ reward for getting to the crafting table and $-1$ for moving into the water and gets no reward for the other features.}
\label{fig:phi}
\end{figure}

MiniCrafter is a game designed to test how SFs handle different scenarios. MiniCrafter was inspired by Crafter~\cite{hafner2021benchmarking}. The game features $5$ object types, $3$ resources, $1$ crafting table and $1$ trap. The agent has an inventory that shows how many resources it has collected from each type. An example state is shown in Figure~\ref{fig:minicrafter} and the state-features are shown with an example task vector in Figure~\ref{fig:phi}. The code for the environment along with the implementation of SFs can be found on Github~\footnote{\url{https://github.com/martinballa/SF-TR}}

The game is grid based $12 \times 12$ with randomly generated levels at each episode. The game is egocentric (agent is always in the centre) and toroidal (the environment is warped, meaning that when the agent moves out on the top it comes back in the bottom). The agent has $4$ actions representing moves in the cardinal directions. The maximum time step an episode can take is $300$.
Moving into a trap ends the episode right away and returns $-1$ as reward (only in the setups with a reward function, during pre-training it ends the episode but the agent does not get the $-1$ reward for it).

To make this environment interesting for transfer learning we set-up a few challenging scenarios in $3$ categories:

\subsubsection{\textbf{Stationary reward function}} \textit{one\_item, two\_item}: The agent has to collect fixed resource type(s) which remains the same throughout the training (wood for \textit{one\_item} and wood and iron for \textit{two\_item}). Picking-up the wrong resource type gives a penalty of $-1$ reward, but does not end the episode. 
\subsubsection{\textbf{Non-stationary reward function}} \textit{random, random\_pen}: A random resource type is sampled at the beginning of each episode. The agent has to get as much of that resource as possible. In the \textit{random} task, if the agent picks up the wrong resource it gets $0$ as reward. In \textit{random\_pen}, if this happens the reward is set to $-1$, as penalty. As the goal changes from episode to episode the agents take the goal vector as input. 
\subsubsection{\textbf{Stationary reward function}} \textit{craft\_staff, craft\_sword, craft\_bow}: In this setting the agent has to collect resources and use them to craft items. To craft a staff the agent only requires wood, for sword it requires wood and iron while for bow it requires all 3 resources. After collecting the required resources, the target item can be crafted by navigating to the crafting table which gives a $+1$ reward. Once the agent gets to the crafting table it disappears and a new table spawns at a randomly chosen location in the environment. Note that the agent only gets the $+1$ reward when it has the required resources for the target item $0$ otherwise. 


These tasks present various challenges: (1) can be presented as a standard RL problem as the target is always stationary. In (2), as the task changes from episode to episode, the agent's policy needs to be conditioned on the goal for the current episode (this is the goal Conditioned RL case). (3) breaks the linearity assumption that SF makes (Equation~\ref{eq:one_step_reward}) since the task cannot be expressed by a single task vector $w$. The problem is that the agent only gets rewarded when it has all the pre-requisite resources and gets to the crafting table. Picking-up the pre-requisites does not give any reward to the agent, so optimising the reward prediction loss can not make the association for collecting the pre-requisite resources.

\subsection{Agents}
In our experiments we have $5$ different ways to train the SF agents. We train DQN and PPO as baselines. The trained agents are indicated in this paper as follows:
\begin{itemize}
    \item SF-1, SF-n: normal SF training method with $1$ and $n$ policies 
    \item SF-HTR-1, SF-HTR-n: SF with Hindsight Task Replacement trained with $1$ and $n$ policies respectively
    \item SF-TR-n: SF with policy objective based task replacement, this method is only available for $n$ policies.
    \item DQN, PPO: standard RL setting except on the non-stationary goal setting where their policies are conditioned on the task vector $w$.
\end{itemize}
Our experiments can be organised into $3$ main categories:
\begin{itemize}
    \item\textbf{Pre-training}: The agent assumes that no reward is provided by the environment ($r(s) = 0)$ for all state $s \in S$. The agent's objective is to acquire knowledge that may be reusable when a target environment is presented.
    \item \textbf{Target training}: Normal RL setup, a reward function is provided and the agent's objective is to learn the policy that maximises the return.
    \item \textbf{Transfer}: After training we evaluate how well the SF agents translate their knowledge to new target tasks without any further training (zero-shot).
\end{itemize}

\subsection{Experimental Setup} \label{ssec:setup}
For baselines we used two standard RL agents: DQN~\cite{mnih2015human}, which is a sample-efficient off-policy agent that predicts the state-action value function (Q-values); and Proximal Policy Optimisation (PPO)~\cite{schulman2017proximal} which is a on-policy, actor-critic method, the actor predicts an action distribution and the critic predicts a value function.

We ran hyper-parameter sweep for all agents. For all agents we tried various learning rates ($1E^{-3}$ to $1E^{-6}$), batch size ($32$ to $256$), replay memory size ($10^5$ to $10^6$) and number of units in the fully connected layers ($32$ to $256$). For PPO we also tuned the replay frequency ($100$ to $4000$) and the number of epochs to optimise on a minibatch ($10$ to $80$). The best found parameters were $0.0001$ for learning rate, for PPO we found that $20$ epochs per update with $4000$ steps between updates worked the best. DQNs were performing similarly, but SF was a bit more sensitive to the learning rate. For the experiments we used $100k$ for replay memory, but increasing it did not seem to make any difference across agents. All the experiments were run on $3$ random seeds. A limit of $5$ million agent-environment interactions per task was used in all experiments. For pre-training, the training interaction budget is equally distributed across tasks.

As some tasks require knowing the inventory counts of the objects we provide agents the inventory in the form of a vector. For the goal conditioned agents the task vector $w$ is also provided (used in \textit{"random"} and \textit{"random\_pen"}). All agents used the same Neural Network architecture. The observations are $12 \times 12 \times 5$ with each dimension representing a different object type. The observation is processed by a convolutional layer with $8$ filters, kernel size $3$ and stride $1$. The inventory and the task vectors in the case of the goal-conditioned tasks are processed by a Fully Connected (FC) layer with $64$ units. The output of the conv layer and the FC layers get concatenated before fed into another FC layer with $64$ units. In the case of DQN and PPO the final layer is another FC layer with $|A|$ outputs. In the case of PPO there is also a head for the critic with a single unit. For SFs there are two final layers, the first with $64$ units while the final layer has $n \times |\phi| \times |A|$ outputs where $n$ is the number of policies used.  

The plots shown in this paper all show the evaluation performance during training. Evaluations were performed after every $20000$ training steps. The shaded areas show the standard error. Table~\ref{tab:transfer} shows the evaluation/transfer performance after the policies have been trained.


\section{Pre-training Results} \label{sec:pretraining}
In the pre-training experiments we are comparing how well the agents learn to maximise each features individually. Note that SF is a single agent while the other agents learn a new set of parameters for each task. The tasks are defined over the feature space $\phi$ in the form of one-hot vectors.

\begin{figure*}[t!]
    \centering
    \subfigure{\includegraphics[width=0.24\textwidth]{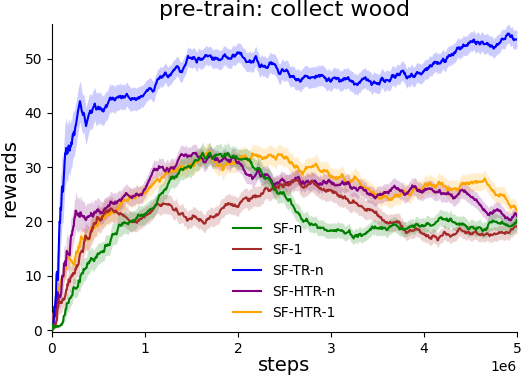}} 
    \subfigure{\includegraphics[width=0.24\textwidth]{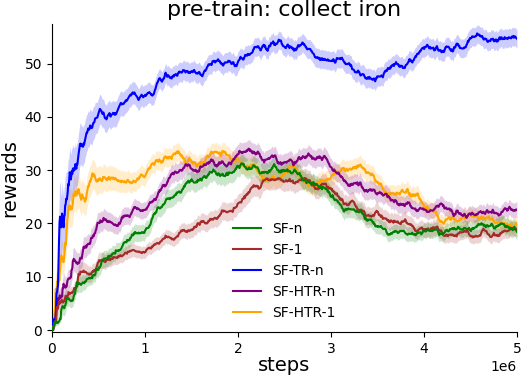}} 
    \subfigure{\includegraphics[width=0.24\textwidth]{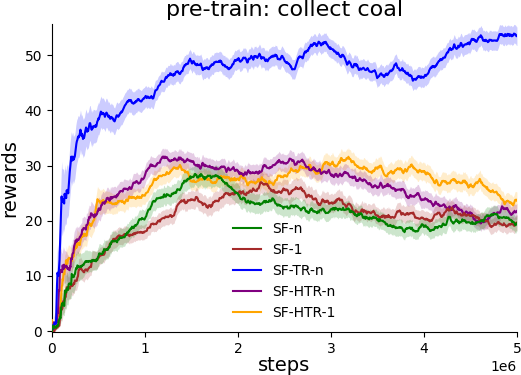}}
    \subfigure{\includegraphics[width=0.24\textwidth]{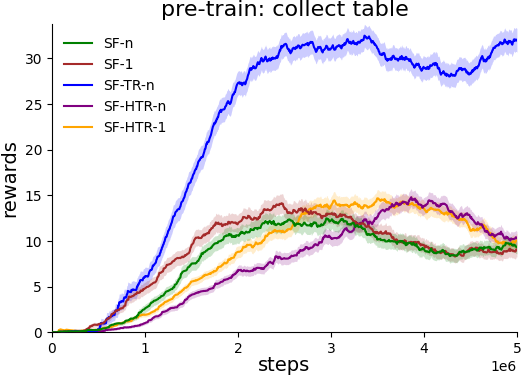}}
    \caption{Comparison of different SF pre-training experiments. HTR means Hindsight Task Replacement, TR means Task Replacement, $n$ and $1$ mean whether we used as many policies as features or a single one. The plots show the cumulative task competition on each task during training: Collect wood, Collect iron, Collect string and Collect table.}
    \label{fig:pretrain_sf}
\end{figure*}

\begin{figure*}[t!]
    \centering
    \subfigure{\includegraphics[width=0.24\textwidth]{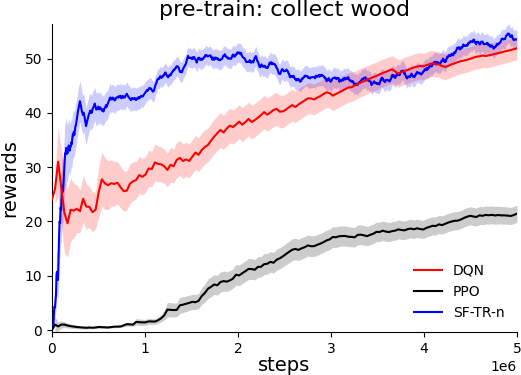}} 
    \subfigure{\includegraphics[width=0.24\textwidth]{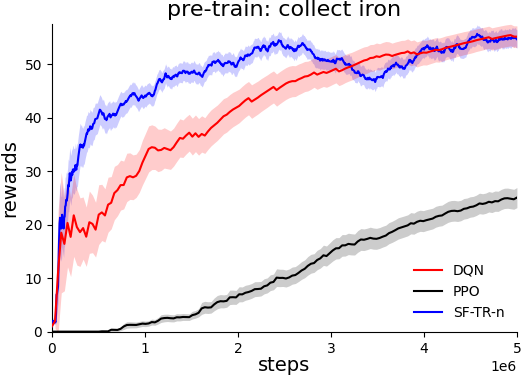}} 
    \subfigure{\includegraphics[width=0.24\textwidth]{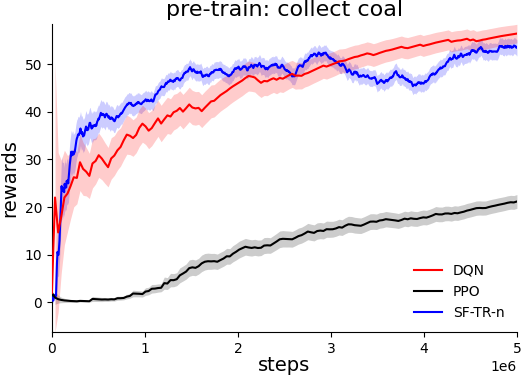}}
    \subfigure{\includegraphics[width=0.24\textwidth]{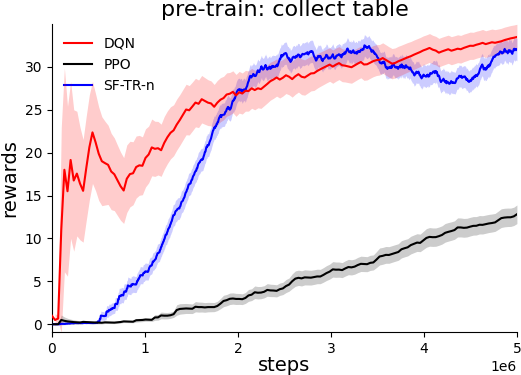}}
    \caption{Pre-training experiments. Comparison of the best SF (Task Replacement with $n$ policies) against baselines. The plots show the cumulative task completion on each task during training: Collect wood, Collect iron, Collect string and Collect table.}
    \label{fig:pretrain_all}
\end{figure*}

The results of the different training methods to train SF are shown in Figure~\ref{fig:pretrain_sf} and comparing the best SF variant to the other baselines is shown in Figure~\ref{fig:pretrain_all}. Note that the final feature was moving into a trap which ended the episode as soon as it was achieved. Since this task was much easier than others we omitted it from the plots (all agents have quickly converged to the maximum reward $1$). During pre-training (Figure~\ref{fig:pretrain_sf}) \textit{SF-TR-n} is noticeably better at maximising each feature individually than the other SF training approaches. Comparing to the baselines (Figure~\ref{fig:pretrain_all}), DQN has a similar final performance as SF, but for pre-training we have to train $n$ separate DQNs while for SF we only have a single network.

\section{Target Training Results} \label{sec:target}
When training on a target environment the environment returns the true reward function for the task. The agent's objective is to get the highest return on the target task. When training SFs on the target environments, we optimised both its parameters $\theta$ and the task vector $w$ simultaneously. For the stationary linear reward functions, SF could learn the exact task vector in less than a $100,000$ interactions (less than $2\%$ of the training) on average when minimising the combined loss function $ L_{SF} + L_{rew}$ (sum of Equation~\ref{eq:sf_loss} and \ref{eq:reward_loss}).

\subsection{Stationary Goal}
\begin{figure*}
\centering
\includegraphics[width=0.28\textwidth]{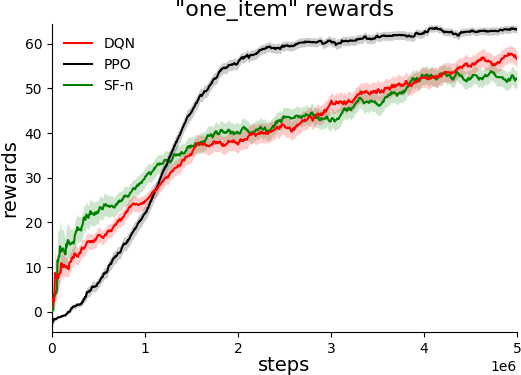}
\includegraphics[width=0.28\textwidth]{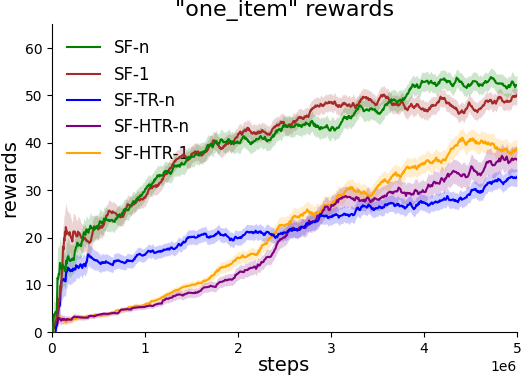}
\includegraphics[width=0.28\textwidth]{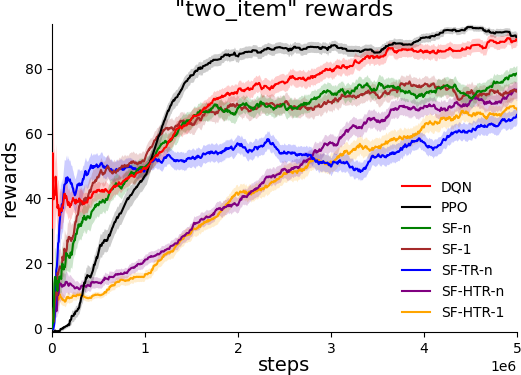}
\centering
\caption{Left side shows the performance of the best SF (standard SF with n policies in this case) compared to the baselines on the task "one\_item". Middle shows the comparison of all SF training methods. Right side shows the comparison of all SF and baselines on the task "two\_item". }
\label{fig:stationary}
\end{figure*}

Standard RL algorithms, SF optimises both its parameters $\theta$ and learns the task vector $w$. Figure~\ref{fig:stationary} shows the results for the tasks $one\_item$ and $two\_item$. 
When training on the target environments directly the benefits of the various SF training methods diminish. In the stationary reward function setting the SF without any task relabelling was the best with both $1$ and $n$ policies, but it was still below PPO's performance. We hypothesise that using Task-relabelling lead to non-optimal policies on this setting as it forces the SF to learn a representation that helps it to solve other tasks not just the one presented as the target task. 

\subsection{Non-stationary Goal}
In this setting a task vector is uniform randomly sampled at the beginning of the episode. Since the agent's objective changes from episode to episode all agents are given the true task vector $w$ as input. Both SF and DQN are goal conditioned in this setting ($a = \pi(s, g)$). SF only learns its parameters $\theta$ and takes the task vector $w$ as input.

\begin{figure*}[t!]
    \centering
    \includegraphics[width=0.24\textwidth]{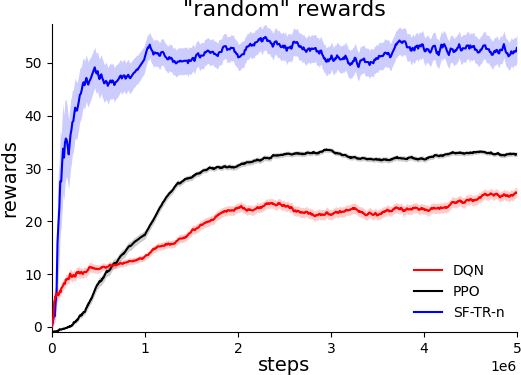}
    \includegraphics[width=0.24\textwidth]{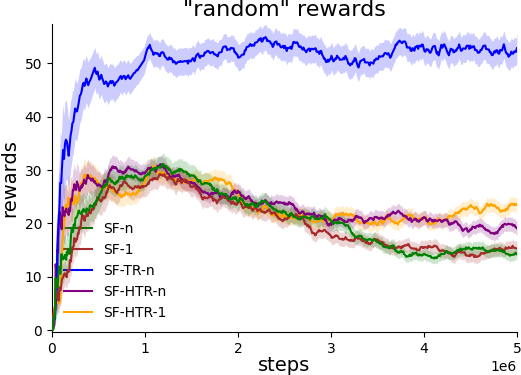}
    \includegraphics[width=0.24\textwidth]{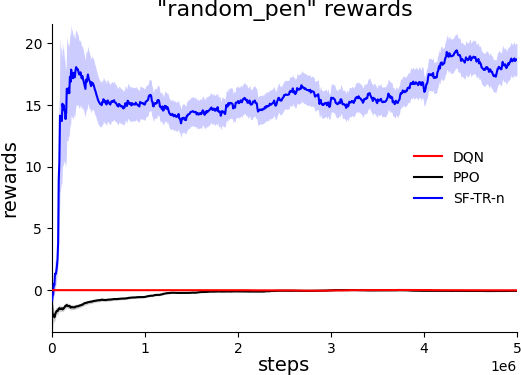}
    \includegraphics[width=0.24\textwidth]{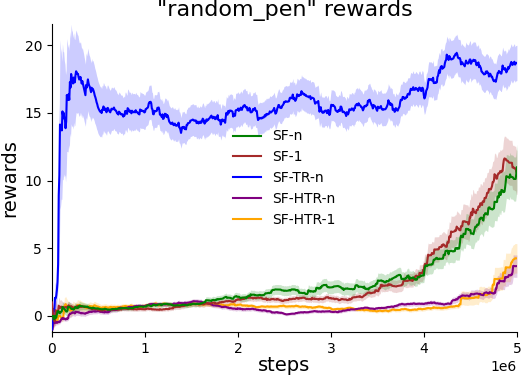}
    \caption{Experiments on the "random" and "random\_pen" targets. The plots show the running mean reward and the standard error (shaded area) during training. From left to right, the first and the third plots show the comparison of the best SF agent against the baseline agents, while the second and the final plots show the comparison of the SF agents with the different training methods used in this paper. }
    \label{fig:random}
\end{figure*}

The results for these experiments are shown in Figure~\ref{fig:random}. SFs were overall much better than the goal conditioned agents. The random\_pen setting was especially challenging for the baselines, we hypothesise that due to the large amount of negative reinforcement and the non-stationary nature of the environment they were unable to learn a policy that would pick-up the correct items instead they learned a policy to avoid all items. Among the SF agents SF-TR-n has quickly learned how to collect the right features and remained stable throughout the training.

\subsection{Stationary non-linear}
The task throughout training does not change, but it is inexpressible by a single task vector $w$, meaning that Equation~\ref{eq:one_step_reward} does not hold. SF learns its parameters $\theta$ and the task vector $w$, other baselines are standard RL algorithms.

\begin{figure*}
\centering
\includegraphics[width=0.24\textwidth]{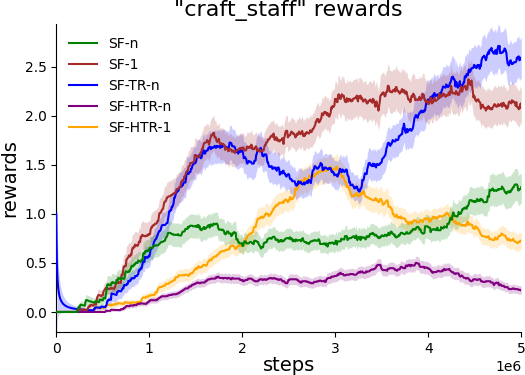}
\includegraphics[width=0.24\textwidth]{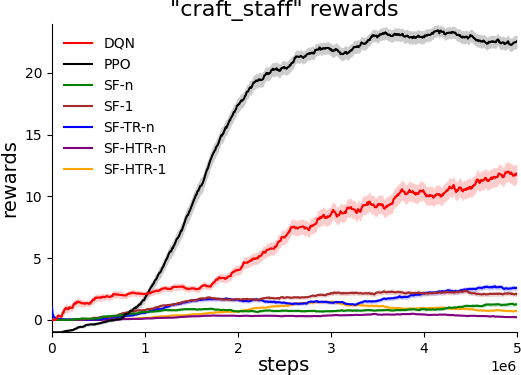}
\includegraphics[width=0.24\textwidth]{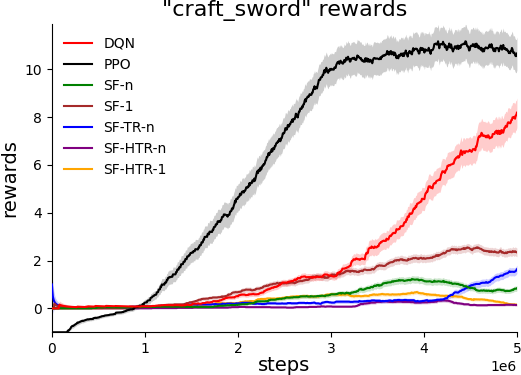}
\includegraphics[width=0.24\textwidth]{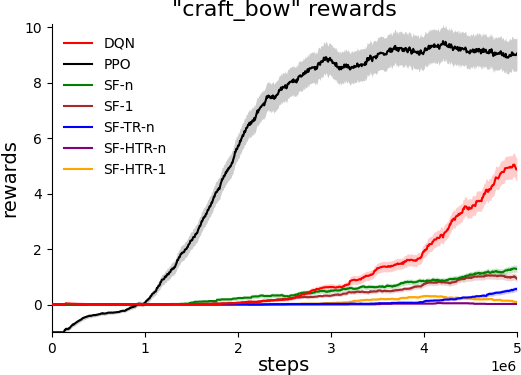}
\caption{Left side shows the comparison of all SF training settings on the task "craft\_staff". The other plots show all SFs compared to the baselines on all the crafting tasks.}
\label{fig:craft}
\end{figure*}

The results in this setting are shown in Figure~\ref{fig:craft}. This setting was designed against SFs, which is clearly visible on the plots. SFs were unable to learn the correct task vectors, which resulted in significantly lower performances than the baseline agents. On the non-linear reward functions (crafting scenarios) we observed that the agent learned very early on that the trap gives a $-1$ penalty, but since picking up the pre-requisite resources did not give any reward it could not bind the rewards with the resources. The agent has managed to learn that getting to the crafting table sometimes give a $+1$ reward, but could not associate it to the resources. An example final vector was $[0, 0, 0, 0.15, -1]$.

\section{Transfer Results} \label{sec:transfer}
Zero-shot transfer: The target task vector $w$ is provided to the agent and the agent's performance is evaluated on the target task without any further training. For the linear reward functions the true task vector is provided, but since it does not exist for the non-linear cases a handcrafted task vector was used. The results of these experiments are shown in Table~\ref{tab:transfer}. We evaluate how SFs transfer from pre-training and also from target environments.

Table~\ref{tab:transfer} shows the evaluation scores in three categories: SF transfer from pre-training, SF transfer from training on a target environment, and baseline agents trained directly on the target environment. 
The transfer results show that agent trained directly on the target environments reach higher scores than the agents transferred from other tasks. The only setting where SFs were clearly better than the standard RL agents was on the non-stationary reward functions where the standard RL agents had to learn how to interpret the task vector. Comparing the transfers from a target environment to other environments we can observe that \textit{SF-TR-n} is the most reliable method. One advantage of the Task relabelling methods is the possibility of transferring to tasks that require features that were unused during training. A good example is $one\_item$ where the agent's objective during training was only to pick-up a single type of resource, so with normal SF it never learns about the other resources or the crafting table. Agents that were trained in the pre-training setup seem to learn more general SFs that helps in transferring, but just giving it the true task vector does not lead to as good performance as if the agent was directly trained on that task. Interestingly, in the case of the non-linear reward functions hand-crafted task vectors were given to the agents which outperformed even the baselines.

We ran a few experiments where we fixed the SF agents parameters $\theta$ after training and only learned the task vector $w$ using the loss function (Equation~\ref{eq:reward_loss}). We noticed that the agent could learn the exact reward function for the stationary linear cases in a few dozen episodes resulting in similar performance to the zero-shot transfer experiments in Table~\ref{tab:transfer}. 



\begin{table*}[t!]
\begin{center}
\begin{tabular}{|l|l|l|l|l|l|l|l|} 
\hline
          & one\_item & two\_item & random & random\_pen & craft\_staff & craft\_sword & craft\_bow \\ \hline
\multicolumn{8}{|c|}{\textbf{pre-train} } \\ \hline
SF-1     & $4.25(0.26)$    & $20.57(1.12)$ & $19.13(1.04)$ & $3.71(0.23)$ & $10.05(0.56)$ & $13.65(0.52)$ & $14.77(0.37)$ \\ \hline
SF-n     & $4.73(0.29)$  & $19.62(1.03)$ & $16.89(0.94)$ & $3.39(0.22)$ & $12.60(0.59)$ & $13.04(0.52)$ & $14.75(0.40)$ \\ \hline
SF-TR-n  & $\textbf{10.63(0.58)}$ & $\textbf{55.89(1.97)}$ & $\textbf{53.48(1.48)}$ & $\textbf{9.10(0.48)}$ & $\textbf{16.76(0.60)}$ & $\textbf{15.01(0.46)}$ & $12.63(0.37)$ \\ \hline
SF-HTR-1 & $5.20(0.41)$  & $24.85(1.62)$ & $22.20(1.47)$ & $4.08(0.34)$ & $10.90(0.68)$ & $14.31(0.64)$ & $15.81(0.45)$ \\ \hline
SF-HTR-n & $5.40(0.30)$  & $26.26(1.24)$ & $23.61(1.06)$ & $3.88(0.23)$ & $12.02(0.51)$ & $14.16(0.45)$ & $15.28(0.35)$ \\ \hline
\multicolumn{8}{|c|}{\textbf{target-train} } \\ \hline
\multicolumn{8}{|l|}{\textbf{one\_item}} \\ \hline
SF-1      & $\textbf{57.91(1.15}$ & $\textbf{70.36(1.16)}$ & $22.11(1.66)$ & $13.67(1.31)$ & $0.40(0.04)$  & $6.47(0.18)$  & $5.40(0.16)$  \\ \hline
SF-n      & $55.53(1.39)$ & $66.36(1.58)$ & $20.14(1.86)$ & $\textbf{19.15(1.71)}$ & $0.41(0.05)$  & $6.89(0.23)$  & $5.72(0.20)$  \\ \hline
SF-TR-n   & $36.46(1.36)$ & $60.54(1.91)$ & $\textbf{44.44(1.48)}$ & $\textbf{18.67(1.21)}$ & $\textbf{10.28(0.39)}$ & $\textbf{10.49(0.41)}$ & $6.14(0.24)$  \\ \hline
SF-HTR-1  & $38.15(1.53)$ & $49.59(1.71)$ & $22.34(1.82)$ & $11.05(1.30)$ & $0.28(0.04)$  & $2.96(0.16)$  & $2.28(0.14)$  \\ \hline
SF-HTR-n  & $38.40(1.76)$ & $57.46(2.02)$ & $17.12(2.00)$ & $12.28(1.64)$ & $0.34(0.06)$  & $3.94(0.20)$  & $3.06(0.19)$  \\ \hline
\multicolumn{8}{|l|}{\textbf{random} }  \\ \hline
SF-1     & $3.18(0.24)$  & $17.58(1.04)$ & $16.49(0.93)$ & $3.72(0.26)$  & $3.09(0.19)$ & $5.01(0.24)$& $7.22(0.26)$  \\ \hline
SF-n     & $2.89(0.19)$  & $16.79(1.03)$ & $15.75(0.91)$ & $3.02(0.19)$  & $2.05(0.15)$  & $4.27(0.23)$  & $6.61(0.26)$  \\ \hline
SF-TR-n  & $\textbf{11.52(0.81)}$ & $\textbf{53.56(2.44)}$ & $\textbf{53.60(1.73)}$ & $\textbf{12.82(0.95)}$ & $\textbf{15.00(0.72)}$ & $\textbf{10.64(0.55)}$ & $\textbf{9.35(0.46)}$  \\ \hline
SF-HTR-1 & $4.79(0.27)$  & $24.23(1.12)$ & $18.70(0.84)$ & $4.70(0.29)$  & $2.59(0.15)$  & $3.87(0.19)$  & $5.77(0.18)$  \\ \hline
SF-HTR-n & $4.17(0.24)$  & $22.32(1.01)$ & $20.21(0.93)$ & $4.84(0.25)$  & $2.39(0.13)$  & $4.61(0.18)$  & $6.39(0.18)$  \\ \hline
\multicolumn{8}{|l|}{\textbf{craft\_staff} } \\ \hline
SF-1         & $1.24(0.09)$  & $3.65(0.23)$  & $2.13(0.15)$  & $-0.05(0.06)$ & $13.23(0.40)$ & $10.75( 0.35)$ & $9.93(0.29)$  \\ \hline
SF-n         & $1.30(0.09)$  & $4.53(0.26)$  & $2.70(0.18)$  & $0.13(0.08)$  & $13.52(0.42)$ & $10.68(0.36)$ & $9.68(0.30)$  \\ \hline
SF-TR-n      & $\textbf{10.14(0.59)}$ & $\textbf{48.06(1.87)}$ & $\textbf{41.71(1.55)}$ & $\textbf{5.84(0.33)}$  & $\textbf{16.26(0.47)}$ & $\textbf{13.97(0.42)}$ & $\textbf{13.23(0.39)}$ \\ \hline
SF-HTR-1     & $1.00(0.08)$  & $2.83(0.19)$  & $2.09(0.15)$  & $-0.08(0.07)$ & $8.37(0.30)$  & $7.76(0.28)$  & $6.86(0.24)$  \\ \hline
SF-HTR-n     & $1.05(0.08)$  & $2.90(0.17)$  & $1.93(0.14)$  & $0.40(0.07)$  & $6.94(0.37)$  & $5.69(0.29)$  & $5.08(0.25)$  \\ \hline
\multicolumn{8}{|c|}{\textbf{baselines} }   \\ \hline
DQN      & $61.14(1.21)$ & $88.20(1.71)$ & $24.34(0.70)$ & $-0.02(0.02)$ & $15.15(0.59)$ & $6.97(0.38)$  & $6.53(0.35)$  \\ \hline
PPO      & $60.65(0.82)$ & $90.16(1.25)$ & $34.86(0.41)$ & $-0.03(0.01)$ & $23.85(0.34)$ & $11.21(0.51)$ & $10.46(0.38)$ \\ \hline
\end{tabular}
\caption{Zero-shot transfer experiments. The labels in the first column show the source the SF agent was trained on while the first row shows the name of the target task. Evaluation is done on the target environments by supplying the true task vector (except in the crafting tasks where a hand-crafted task vector was supplied). The numbers in the cells represent the mean and the standard error over $100$ evaluation episodes for each seed. The results with the highest values are in bold font.  The table is separated into blocks based on the source task used for training. The baselines (DQN and PPO) were trained directly on the target environment (no transfer just evaluation).}
\label{tab:transfer}
\end{center}
\end{table*}



\section{Conclusion and Future Work} \label{sec:conclusion}
In this paper we have presented $2$ relabelling methods that could be used to train SFs for better performance. Unfortunately, Hindsight Task Relabelling (HTR) did not show a great improvement, but Task Replacement (TR) with $n$ policies has shown much better performance for both training and transferring in most settings. We compared the benefits of using $1$ vs $n$ policies and got to the conclusion that using $1$ policy rarely results in better performance. We have investigated how the training task for SF results in transfer to tasks across $3$ types of target reward functions. Overall, our proposed agent \textit{SF-TR-n} performs the best among the SF agents. Although they do not get better results than standard RL agents trained specifically on the target tasks, the biggest advantage of SFs is that they can easily transfer to new tasks without any further training (or little training if task vector needs to be learned), while for the baseline agents we need to do a full training for each variant.

SFs have great potential to be used when various behaviours are required from a single agent. In this work one of the biggest limitations was the non-linear reward function. To overcome this limitation we may redesign the state features $\phi(s)$ in a way that Equation~\ref{eq:one_step_reward} holds (it remains linear), in the crafting environments defining the feature when the agent gets to the crafting table to only return $1$ when it has all the pre-requisites could solve this issue, but it requires engineering all features that may come up in the target problems.
If the optimal policy is not required, reward shaping after training is also an option by manually crafting task vectors and evaluation the agent on them which could be enough in some cases (for example automated playtesting). 


As future work using the training methods presented in this work could be combined with Hierarchical RL such as the Option-Keyboard~\cite{barreto2019option} setup which could facilitate transferring to non-linear target reward functions as we could alternate between different task vectors. Another improvement could be during the pre-training phase, when task vectors are uniformly selected at random. This procedure could be replaced by a method that selects task vectors with the right difficulty based on the agent's learning progress. 


\section*{Acknowledgments}
This work was funded by the EPSRC Centre for Doctoral Training in Intelligent Games and Game Intelligence (IGGI) EP/L015846/1.
 
\bibliography{ref}

\begin{thebibliography}{10}
\providecommand{\url}[1]{#1}
\csname url@samestyle\endcsname
\providecommand{\newblock}{\relax}
\providecommand{\bibinfo}[2]{#2}
\providecommand{\BIBentrySTDinterwordspacing}{\spaceskip=0pt\relax}
\providecommand{\BIBentryALTinterwordstretchfactor}{4}
\providecommand{\BIBentryALTinterwordspacing}{\spaceskip=\fontdimen2\font plus
\BIBentryALTinterwordstretchfactor\fontdimen3\font minus
  \fontdimen4\font\relax}
\providecommand{\BIBforeignlanguage}[2]{{%
\expandafter\ifx\csname l@#1\endcsname\relax
\typeout{** WARNING: IEEEtran.bst: No hyphenation pattern has been}%
\typeout{** loaded for the language `#1'. Using the pattern for}%
\typeout{** the default language instead.}%
\else
\language=\csname l@#1\endcsname
\fi
#2}}
\providecommand{\BIBdecl}{\relax}
\BIBdecl

\bibitem{mnih2015human}
V.~Mnih, K.~Kavukcuoglu, D.~Silver, A.~A. Rusu, J.~Veness, M.~G. Bellemare,
  A.~Graves, M.~Riedmiller, A.~K. Fidjeland, G.~Ostrovski \emph{et~al.},
  ``Human-level control through deep reinforcement learning,'' \emph{Nature},
  vol. 518, no. 7540, p. 529, 2015.

\bibitem{silver2016mastering}
D.~Silver, A.~Huang, C.~J. Maddison, A.~Guez, L.~Sifre, G.~Van Den~Driessche,
  J.~Schrittwieser, I.~Antonoglou, V.~Panneershelvam, M.~Lanctot \emph{et~al.},
  ``Mastering the game of go with deep neural networks and tree search,''
  \emph{nature}, vol. 529, no. 7587, pp. 484--489, 2016.

\bibitem{vinyals2019grandmaster}
O.~Vinyals, I.~Babuschkin, W.~M. Czarnecki, M.~Mathieu, A.~Dudzik, J.~Chung,
  D.~H. Choi, R.~Powell, T.~Ewalds, P.~Georgiev \emph{et~al.}, ``Grandmaster
  level in starcraft ii using multi-agent reinforcement learning,''
  \emph{Nature}, vol. 575, no. 7782, pp. 350--354, 2019.

\bibitem{kirk2021survey}
R.~Kirk, A.~Zhang, E.~Grefenstette, and T.~Rockt{\"a}schel, ``A survey of
  generalisation in deep reinforcement learning,'' \emph{arXiv preprint
  arXiv:2111.09794}, 2021.

\bibitem{barreto2016successor}
A.~Barreto, W.~Dabney, R.~Munos, J.~J. Hunt, T.~Schaul, H.~Van~Hasselt, and
  D.~Silver, ``Successor features for transfer in reinforcement learning,''
  \emph{arXiv preprint arXiv:1606.05312}, 2016.

\bibitem{bergdahl2020augmenting}
J.~Bergdahl, C.~Gordillo, K.~Tollmar, and L.~Gissl{\'e}n, ``Augmenting
  automated game testing with deep reinforcement learning,'' in \emph{2020 IEEE
  Conference on Games (CoG)}, 2020, pp. 600--603.

\bibitem{guerrero2018using}
C.~Guerrero-Romero, S.~M. Lucas, and D.~Perez-Liebana, ``Using a team of
  general ai algorithms to assist game design and testing,'' in \emph{IEEE
  Conf. on Computational Intelligence and Games (CIG)}, 2018, pp. 1--8.

\bibitem{johnson2016malmo}
M.~Johnson, K.~Hofmann, T.~Hutton, and D.~Bignell, ``The malmo platform for
  artificial intelligence experimentation.'' in \emph{IJCAI}.\hskip 1em plus
  0.5em minus 0.4em\relax Citeseer, 2016, pp. 4246--4247.

\bibitem{sutton2018reinforcement}
R.~S. Sutton and A.~G. Barto, \emph{Reinforcement learning: An
  introduction}.\hskip 1em plus 0.5em minus 0.4em\relax MIT press, 2018.

\bibitem{dayan_improving_1993}
\BIBentryALTinterwordspacing
P.~Dayan, ``\BIBforeignlanguage{en}{Improving {Generalization} for {Temporal}
  {Difference} {Learning}: {The} {Successor} {Representation}},''
  \emph{\BIBforeignlanguage{en}{Neural Computation}}, vol.~5, no.~4, pp.
  613--624, Jul. 1993. [Online]. Available:
  \url{https://www.mitpressjournals.org/doi/abs/10.1162/neco.1993.5.4.613}
\BIBentrySTDinterwordspacing

\bibitem{kulkarni2016deep}
T.~D. Kulkarni, A.~Saeedi, S.~Gautam, and S.~J. Gershman, ``Deep successor
  reinforcement learning,'' 2016.

\bibitem{barreto2019option}
A.~Barreto, D.~Borsa, S.~Hou, G.~Comanici, E.~Ayg{\"u}n, P.~Hamel, D.~Toyama,
  J.~Hunt, S.~Mourad, D.~Silver \emph{et~al.}, ``The option keyboard combining
  skills in reinforcement learning,'' in \emph{Proceedings of the 33rd
  International Conference on Neural Information Processing Systems}, 2019, pp.
  13\,052--13\,062.

\bibitem{hansen2020fast}
S.~Hansen, W.~Dabney, A.~Barreto, T.~V. de~Wiele, D.~Warde-Farley, and V.~Mnih,
  ``Fast task inference with variational intrinsic successor features,'' 2020.

\bibitem{filos2021psiphi}
A.~Filos, C.~Lyle, Y.~Gal, S.~Levine, N.~Jaques, and G.~Farquhar,
  ``Psiphi-learning: Reinforcement learning with demonstrations using successor
  features and inverse temporal difference learning,'' \emph{arXiv preprint
  arXiv:2102.12560}, 2021.

\bibitem{machado2020count}
M.~C. Machado, M.~G. Bellemare, and M.~Bowling, ``Count-based exploration with
  the successor representation,'' in \emph{Proceedings of the AAAI Conference},
  vol.~34, no.~04, 2020, pp. 5125--5133.

\bibitem{momennejad2017successor}
I.~Momennejad, E.~M. Russek, J.~H. Cheong, M.~M. Botvinick, N.~D. Daw, and
  S.~J. Gershman, ``The successor representation in human reinforcement
  learning,'' \emph{Nature human behaviour}, vol.~1, no.~9, pp. 680--692, 2017.

\bibitem{schaul2015universal}
T.~Schaul, D.~Horgan, K.~Gregor, and D.~Silver, ``Universal value function
  approximators,'' in \emph{International Conference on Machine Learning},
  2015, pp. 1312--1320.

\bibitem{andrychowicz2017hindsight}
M.~Andrychowicz, F.~Wolski, A.~Ray, J.~Schneider, R.~Fong, P.~Welinder,
  B.~McGrew, J.~Tobin, O.~P. Abbeel, and W.~Zaremba, ``Hindsight experience
  replay,'' in \emph{Advances in Neural Information Processing Systems}, 2017,
  pp. 5048--5058.

\bibitem{borsa2018universal}
D.~Borsa, A.~Barreto, J.~Quan, D.~Mankowitz, R.~Munos, H.~van Hasselt,
  D.~Silver, and T.~Schaul, ``Universal successor features approximators,''
  \emph{arXiv:1812.07626}, 2018.

\bibitem{barreto2020fast}
A.~Barreto, S.~Hou, D.~Borsa, D.~Silver, and D.~Precup, ``Fast reinforcement
  learning with generalized policy updates,'' \emph{Proceedings of the National
  Academy of Sciences}, vol. 117:48, pp. 30\,079--30\,087, 2020.

\bibitem{hafner2021benchmarking}
\BIBentryALTinterwordspacing
D.~Hafner, ``Benchmarking the spectrum of agent capabilities,'' in \emph{Deep
  RL Workshop NeurIPS 2021}, 2021. [Online]. Available:
  \url{https://openreview.net/forum?id=76iypkcozLU}
\BIBentrySTDinterwordspacing

\bibitem{schulman2017proximal}
J.~Schulman, F.~Wolski, P.~Dhariwal, A.~Radford, and O.~Klimov, ``Proximal
  policy optimization algorithms,'' \emph{arXiv:1707.06347}, 2017.

\end{thebibliography}
\bibliographystyle{IEEEtran}

\vfill

\end{document}